\documentclass{article}
\usepackage{spconf,amsmath,graphicx}

\usepackage{url}
\usepackage{tabularx}
\usepackage[ruled,vlined]{algorithm2e}
\usepackage{booktabs}
\usepackage{multirow}
\usepackage[numbers]{natbib}
\usepackage{subcaption}

\title{Not All Minorities Are Equal: Fairness Outcomes in Private Image Generation}
\name{Cohen Archbold$^{1}$, Usman Hassan$^{2}$, Sen-ching Samson Cheung$^{2}$}

\address{
$^{1}$Department of Computer Science, University of Kentucky\\
$^{2}$Department of Electrical and Computer Engineering, University of Kentucky
}

\begin{document}
%
\maketitle
\begin{abstract}
Differential privacy (DP) provides formal guarantees for protecting individuals in artificial intelligence systems, but its consequences for fairness in generative models remain poorly understood. We investigate how privacy-preserving training affects the representation of demographic groups and visual attributes in generative models of human faces. Across two facial image datasets and multiple privacy levels, we find that stronger privacy can disproportionately degrade the quality and utility of some underrepresented groups, suggesting that privacy can amplify existing disparities. However, representation alone does not explain these effects: several well-represented attributes also experience substantial degradation, while some underrepresented attributes remain comparatively robust. These findings show that the impact of privacy-preserving learning cannot be explained solely by data representation and may depend on the characteristics and interactions of the information being learned. Our results highlight the need to consider both privacy and fairness when evaluating generative models.

\end{abstract}
\section{Introduction}
Artificial intelligence (AI) is increasingly applied to domains involving sensitive visual data, including medical imaging, biometrics, and social media analysis. The use of such data raises concerns about demographic bias~\cite{karkkainen2021fairface}, identity leakage and potential misuse~\cite{carlini2023extracting}. In these settings, models must balance the utility of learned representations with the need to protect individuals whose data are used for training. This challenge is particularly important for generative models, as their ability to generate and share synthetic data introduces the possibility that information memorized during training may be revealed through generated samples, while demographic groups may be represented unequally. Differential Privacy (DP)~\cite{Dwork2006OurDO}  provides a principled framework for limiting the influence of individual training examples and protecting against information leakage from model parameters and outputs. However, stronger privacy typically reduces model performance, image quality, and fidelity to the underlying data distribution~\cite{lyu2024differentially,kim2025pfguard}. For generative models, this degradation may not occur uniformly across different subsets of the data, potentially introducing disparities between demographic groups.

Prior work~\cite{kamath2025bias} has theorized and empirically observed that differential privacy can disproportionately suppress the contributions of minority groups during training, with such effects reported in language models~\cite{bagdasaryan2019differential,tenorio2026doesdifferentialprivacyaffect}, tabular data~\cite{bagdasaryan2019differential,bullwinkel2022evaluatingfairnessdp,ganev2022robin}, and small image datasets~\cite{bagdasaryan2019differential,esipova2023disparate,ganev2022robin}. Several studies have also explored dataset re-balancing as a means of mitigating these disparities in DP generative models~\cite{xu2021ffpdg,kim2025pfguard}. In the context of human faces, prior work has primarily examined these effects through binary downstream classification tasks, such as gender classification, linking class prevalence to degradation in classification accuracy~\cite{bagdasaryan2019differential,bullwinkel2022evaluatingfairnessdp}. Whether similar relationships emerge across multiple demographic subgroups and semantic attributes in DP generative models remains less understood. We therefore investigate this relationship in Differentially Private Latent Diffusion Models (DP-LDMs)~\cite{lyu2024differentially}, evaluating generated images across demographic subgroups and semantic attributes under multiple privacy budgets. We operationalize fairness as equality of degradation relative to the corresponding non-private baseline across demographic/semantic subgroups. We find that underrepresented groups can indeed experience disproportionate degradation, but this relationship is not universal. Several relatively well-represented attributes also exhibit substantial degradation, suggesting that spatial context, attribute interactions, and characteristics of individual subsets may influence how privacy-preserving training affects generated images.

The primary contributions of this work are summarized as follows:

\begin{enumerate}
\item We provide an empirical evaluation of demographic fairness in Latent Diffusion Models trained under Differential Privacy constraints.

\item We investigate the relationship between attribute prevalence and degradation in generated image quality and utility across varying privacy budgets.

\item We demonstrate that demographic representation alone does not fully explain fairness outcomes under DP, identifying substantial variation across demographic groups and visual attributes.

\end{enumerate}

\section{Background}

\textbf{Differential Privacy (DP)}
provides formal privacy guarantees by limiting how much a single record can influence a mechanism’s output~\cite{Dwork2006OurDO}. Let $X,X'\in\mathcal{X}^n$ be neighboring datasets differing in one data point. A randomized mechanism $\mathcal{M}$ satisfies $(\varepsilon,\delta)$-Differential Privacy if, for all measurable sets $S$,
\begin{equation}
\label{eqappDP}
\mathrm{P}(\mathcal{M}(X)\in S)
\leq
e^{\varepsilon}\mathrm{P}(\mathcal{M}(X')\in S)+\delta .
\end{equation}
Here, $\varepsilon$ controls the strength of the privacy guarantee, with smaller values corresponding to stronger privacy. $\delta$ is a small additive slack parameter that relaxes the pure $\epsilon$-DP guarantee. This mechanism has been applied in deep learning across many domains, enabled by algorithms such as DP-SGD~\cite{abadi2016deep} and various frameworks such as PATE~\cite{papernot2018scalable}.

\noindent
\textbf{Differentially Private Latent Diffusion Models (DP-LDMs)}
build upon latent diffusion models (LDMs), which perform the diffusion process in a lower-dimensional latent space to reduce computational cost and privacy expenditure while maintaining high-quality generation~\cite{rombach2022high}. DP-LDMs leverage models pre-trained on large public datasets and subsequently fine-tune them on private data using DP-SGD~\cite{abadi2016deep}. During private fine-tuning, the model weights are frozen except for the attention layers, which are updated using differentially private gradients. This allows the model to retain knowledge learned from public data while adapting to the private dataset under formal privacy guarantees.


\section{Methodology}

\begin{figure}[t]
    \centering

    \begin{subfigure}{\columnwidth}
        \centering
        \includegraphics[width=\columnwidth]{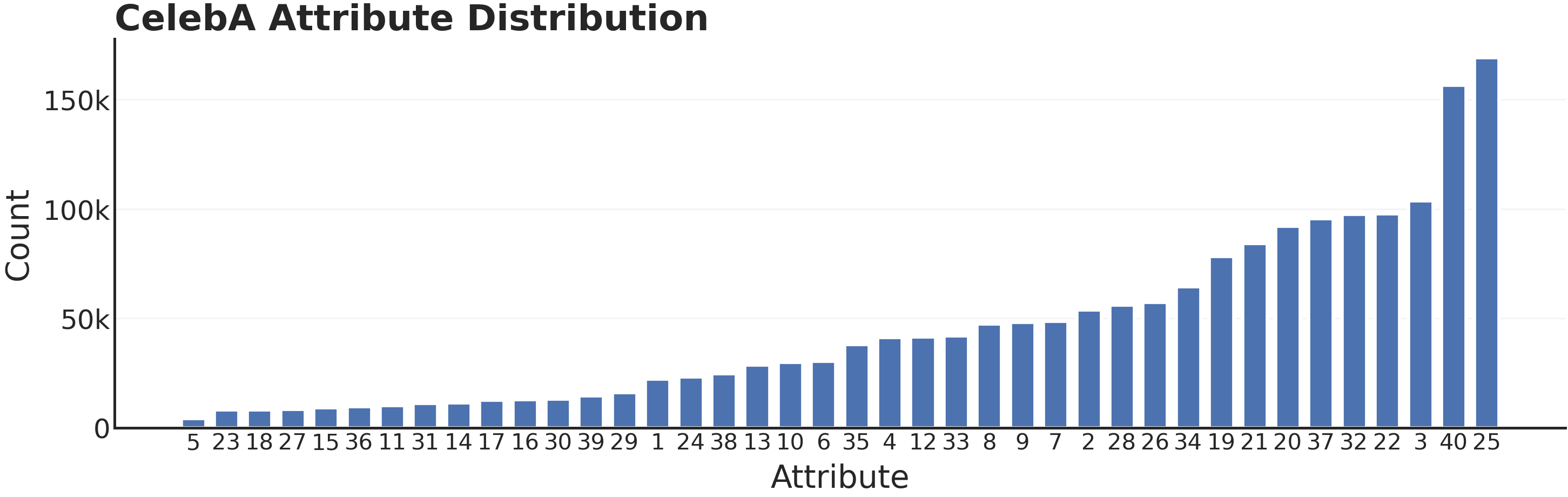}
        \label{fig:celeba_distribution}
    \end{subfigure}

    \vspace{0.5em}

    \begin{subfigure}{\columnwidth}
        \centering
        \includegraphics[width=\columnwidth]{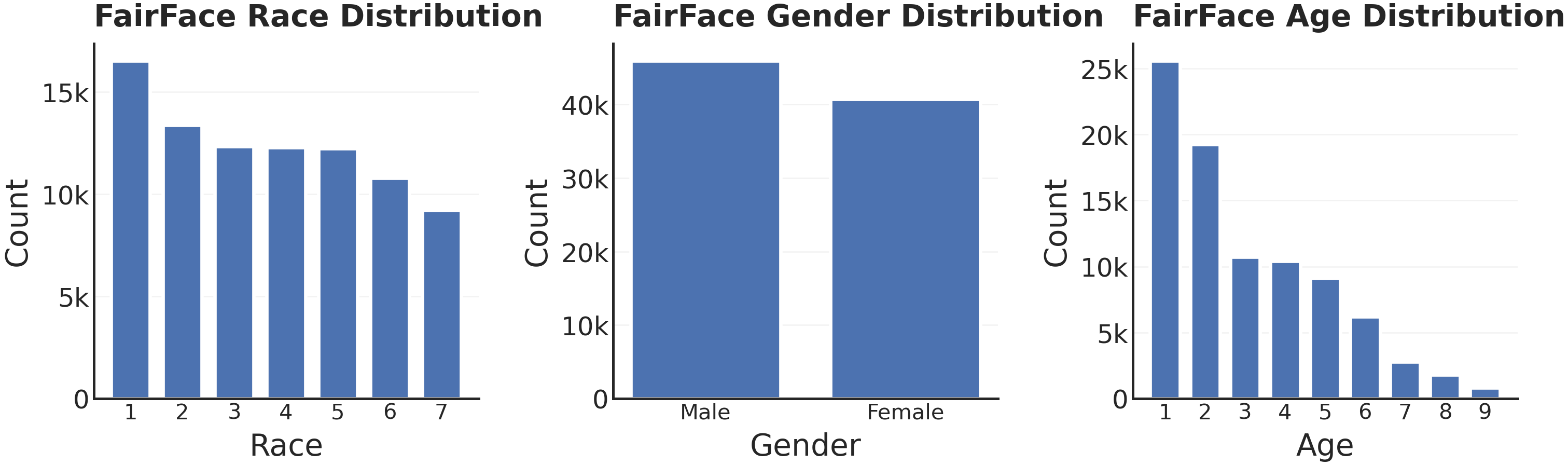}
        \label{fig:fairface_distribution}
    \end{subfigure}

    \caption{Comparison of demographic and attribute distributions in the CelebA and FairFace datasets. The categories are ordered by their frequency of occurrence.}
    \label{fig:dataset_distributions}
\end{figure}

We use three facial-image datasets with distinct roles. Microsoft Synth~\cite{wood2021fake}, a publicly available dataset of fully 3D-rendered faces, is used for pretraining. CelebA~\cite{liu2015faceattributes} and FairFace~\cite{karkkainen2021fairface} serve as the two target datasets for differentially private fine-tuning and evaluation. CelebA provides 40 facial attributes with substantial variation in attribute prevalence, whereas FairFace provides demographic annotations for race, gender, and age. Figure~\ref{fig:dataset_distributions} compares the distributions of attributes and demographic groups in the two target datasets, and Table~\ref{tab:datasets} summarizes all three datasets used in the study.

\begin{table}[h]
    \centering
    \scriptsize
    \setlength{\tabcolsep}{3pt}
    \begin{tabularx}{\columnwidth}{l c c X c}
        \hline
        \textbf{Dataset} &
        \textbf{\# Samples} &
        \textbf{Resolution} &
        \textbf{Labels} &
        \textbf{\# IDs} \\
        \hline

        Microsoft Synth~\cite{wood2021fake} &
        100,000 &
        $512{\times}512$ &
        -- & 
        -- \\

        CelebA~\cite{liu2015faceattributes} &
        202,599 &
        $178{\times}218$ &
        40 attributes: gender, age, hair color, facial hair, eyeglasses, etc. &
        10,177 \\

        FairFace~\cite{karkkainen2021fairface} &
        108,501 &
        $224{\times}224$ &
        Race (7), gender (2), age (9) &
        -- \\
        
    \end{tabularx}
    \caption{Comparison of datasets used in this work.}
    \label{tab:datasets}
\end{table}

\subsection{Fine-Tuned Conditional DP-LDM}

We use conditional Latent Diffusion Models (LDMs) to generate facial images conditioned on the multi-label annotations associated with each training example. Following unconditional pretraining, we independently fine-tune the models on CelebA and FairFace using either standard optimization (No-DP) or differentially private optimization with $\epsilon=10$ or $\epsilon=1$. Following the original DP-LDM formulation, fine-tuning updates only the conditioning (cross-attention) layers of the UNet, while the remaining UNet parameters and the latent encoder remain frozen. Thus, differential privacy is applied to the trainable conditioning layers during adaptation to the target dataset. For each target dataset, all privacy settings use the same pretrained initialization, enabling controlled comparison of the effect of the privacy constraint. To sample datasets for downstream training, we use the original labels for each dataset as conditions, ensuring consistent label composition across privacy settings and allowing us to isolate differences in downstream performance attributable to the generated data.

\begin{figure*}[h]
    \centering
    \includegraphics[width=\linewidth]{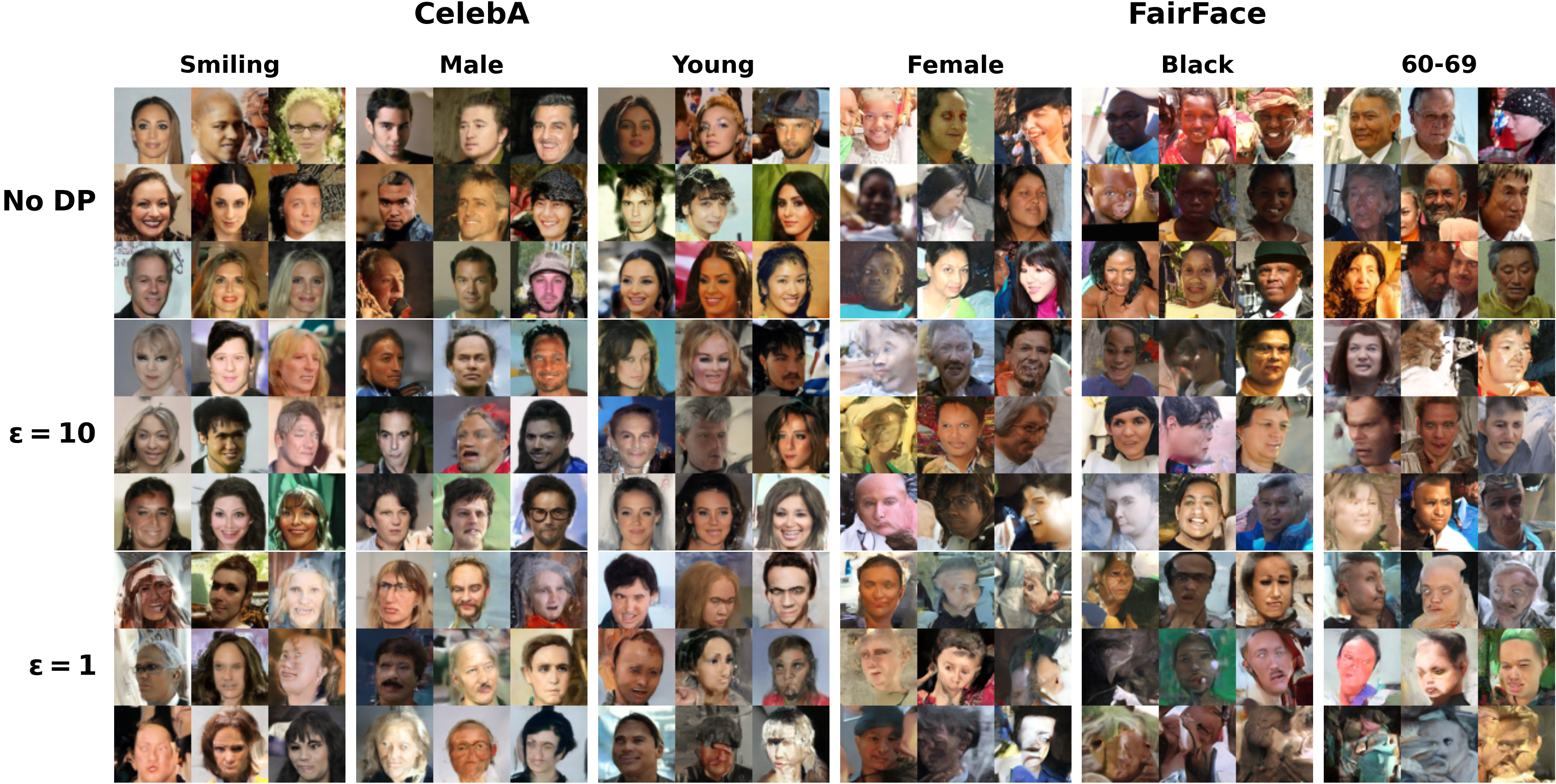}
    \caption{Qualitative comparison of images generated with differentially private (DP) training across privacy budgets. The three left columns show samples conditioned on CelebA attributes (Smiling, Male, and Young), while the three right columns show samples conditioned on FairFace attributes (Female, Black, and 60–69). The results illustrate the effect of increasing privacy constraints on conditional image generation and attribute fidelity.}
    \label{fig:placeholder}
\end{figure*}

\subsection{Downstream Evaluation}
We evaluate generated data using downstream multi-attribute classification. For each privacy setting, we train multiple classifier architectures using the corresponding synthetic training dataset and evaluate them on a held-out real-data test set. For CelebA, we use a single multi-label classification head to jointly predict the target attributes. For FairFace, we use three task-specific classification heads, with race and age group formulated as multi-class classification tasks over their respective categories. To account for class imbalance, we apply positive class reweighting during training. Results are reported separately for each attribute, allowing us to identify degradation that may be obscured by aggregate performance. Standard deviations are computed across five classifier-training runs with different random initializations. We also calculate Fr\'echet Inception Distance (FID)~\cite{heusel2018gans}, by computing the InceptionV3~\cite{szegedy2016rethinking} features of both synthetic and real image datasets.


\begin{table}
    \centering
    \caption{
        Quality and downstream utility performance on
        CelebA and FairFace under different privacy budgets.
        $\downarrow$ indicates lower is better and $\uparrow$ indicates
        higher is better.
    }
    \label{tab:main_results}
    \resizebox{\linewidth}{!}{%
    \begin{tabular}{llccc}
        \toprule
        \textbf{Dataset} & \textbf{Metric}
        & \textbf{No DP}
        & \boldmath$\epsilon=10$
        & \boldmath$\epsilon=1$ \\
        \midrule

        \multirow{4}{*}{CelebA}
        & FID $\downarrow$
        & 10.11 & 15.93 & 38.67 \\

        & Accuracy $\uparrow$
        & $	0.853 \pm 0.001$ & $0.794 \pm 0.003 $ & $0.693 \pm 0.004 $ \\

        & Macro-F1 $\uparrow$
        & $	0.605 \pm 0.002$ & $0.459 \pm 0.005$ & $0.267 \pm 0.012$ \\

        & ROC-AUC $\uparrow$
        & $0.891 \pm 0.001$ & $0.789 \pm 0.004$ & $0.548 \pm 0.012$ \\

        \midrule

        \multirow{4}{*}{FairFace}
        & FID $\downarrow$
        & 12.23 & 31.23 & 62.44 \\

        & Accuracy $\uparrow$
        & $	0.593 \pm 0.003$ & $0.344 \pm 0.007$ & $0.279 \pm 0.004$ \\

        & Macro-F1 $\uparrow$
        & $0.581 \pm 0.002$ & $	0.309 \pm 0.007$ & $0.251 \pm 0.004$ \\

        & ROC-AUC $\uparrow$
        & $0.872 \pm 0.002$ & $	0.595 \pm 0.005$ & $0.514 \pm 0.004$ \\

        \bottomrule
    \end{tabular}
    }
\end{table}

\noindent
\textbf{Implementation Details:} All images are resized to a spatial resolution of $64\times64$. For each target dataset, models are trained under three privacy settings: a non-private baseline (No-DP), $\varepsilon=10$, and $\varepsilon=1$, where $\varepsilon=1$ provides the stronger privacy guarantee. All other training, sampling, and evaluation procedures are kept fixed across privacy settings to isolate the effect of differential privacy. Experiments were conducted using 4 NVIDIA A100s with 48 GB of VRAM using PyTorch.\footnote{\textbf{Acknowledgment:} This work used Delta advanced computing and data resource at University of Illinois Urbana-Champaign through allocation CIS230383 from the Advanced Cyberinfrastructure Coordination Ecosystem: Services \& Support (ACCESS) program, which is supported by U.S. National Science Foundation grants \#2138259, \#2138286, \#2138307, \#2137603, and \#2138296.} Models were trained for 150 epochs with a batch size of 8192. For DP training, we use the Gaussian Differential Privacy~\cite{10.1111/rssb.12454} (GDP) accountant with  a clipping norm of 0.0005 and $\delta = 1\times 10^{-6}$. For downstream classification, we trained a ResNet-18 using AdamW with a learning rate of $3\times 10^{-4}$, weight decay of $1\times 10^{-4}$, batch size of 128, and trained for 30 epochs.

\section{Results \& Discussion}





\begin{figure*}[!h]
    \centering

    \begin{subfigure}{0.46\textwidth}
        \centering
        \includegraphics[width=\linewidth]{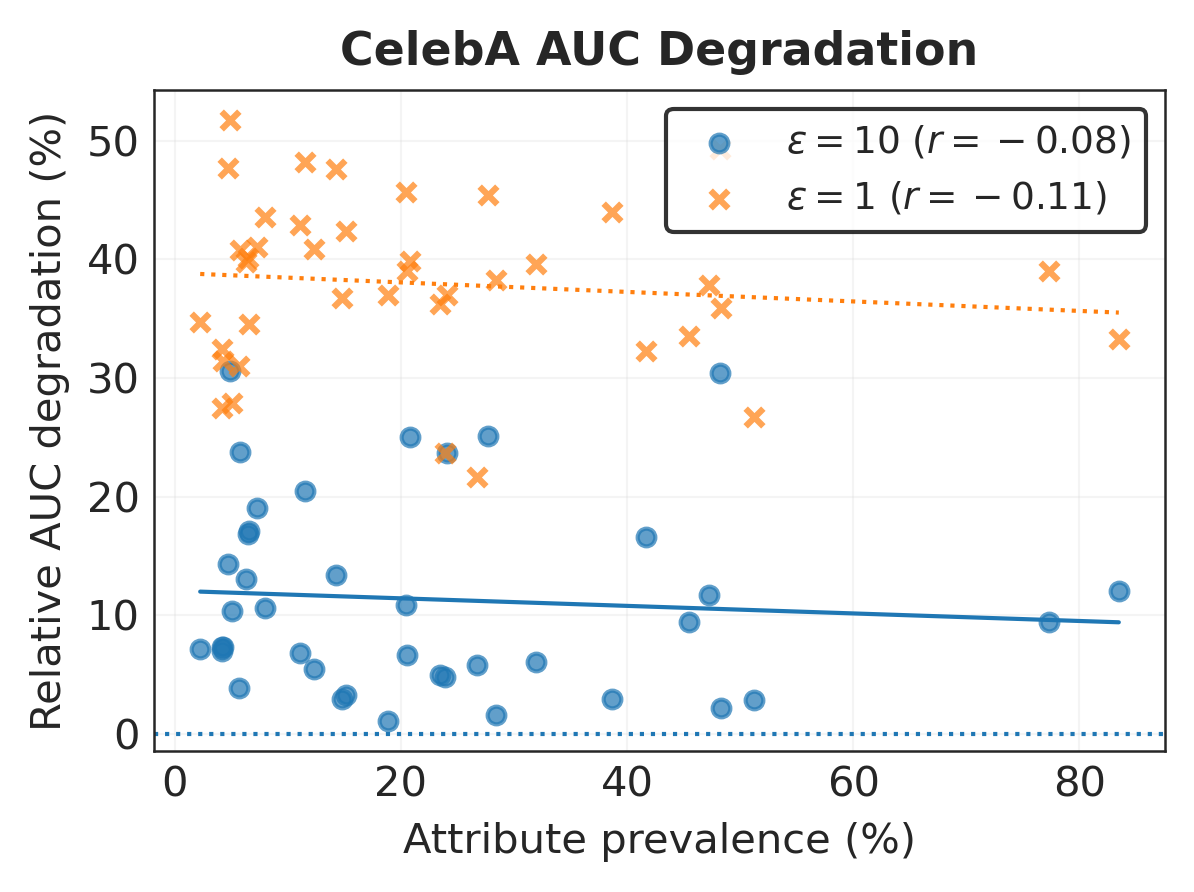}
        \label{fig:celeba_auc}
    \end{subfigure}
    \hfill
    \begin{subfigure}{0.46\textwidth}
        \centering
        \includegraphics[width=\linewidth]{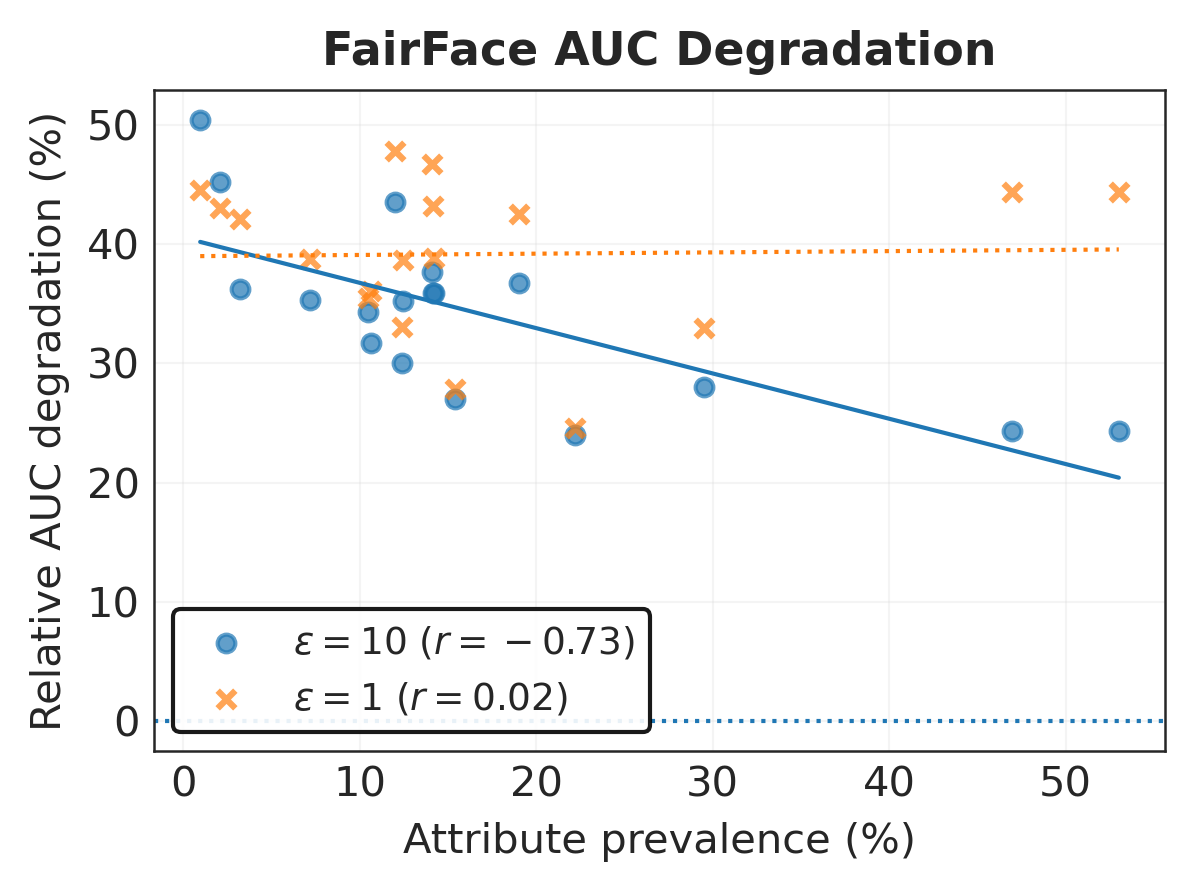}
        \label{fig:fairface_auc}
    \end{subfigure}

    \caption{Relative attribute performance degradation in AUC versus prevalence for CelebA (left) and FairFace (right). Degradation is computed relative to the non-private model as $(AUC_{nodp}-AUC_{\epsilon})/AUC_{nodp}\times100\%$, such that positive values indicate worse performance. Points denote attributes while lines show linear fits and corresponding Pearson correlations ($r$).}
    \label{fig:auc_degradation}
\end{figure*}

\subsection{Overall Downstream Utility}

We report overall FID, Accuracy, Macro-F1, and ROC-AUC of downstream ResNet18 models trained on synthetic datasets produced by DP-LDM at various privacy levels in Table \ref{tab:main_results}. As expected, overall performance degrades as $\varepsilon$ decreases across both datasets. Notably at $\varepsilon=1$, downstream classifier performance substantially drops, representing significant degradation in attribute modeling. As shown in Fig \ref{fig:placeholder}, we find that DP-LDM has much greater difficulty modeling the FairFace distribution, likely due to the variety in head pose, background and lighting that FairFace exhibits over CelebA.


\begin{table}
    \centering
    \caption{
        Relationship between attribute prevalence and relative AUC degradation
        under differential privacy. AUC degradation is computed relative to the
        corresponding No-DP baseline. Pearson $r$, Spearman $\rho$, and
        permutation-test $p$-values quantify the association between attribute
        prevalence and AUC degradation. $p_{\mathrm{FDR}}$ denotes the
        Benjamini--Hochberg FDR-corrected permutation-test $p$-value.
    }
    \label{tab:auc_degradation_statistics}
    \resizebox{\linewidth}{!}{%
    \begin{tabular}{llcccc}
        \toprule
        \textbf{Dataset}
        & \boldmath$\epsilon$
        & \textbf{Pearson $r$}
        & \textbf{Spearman $\rho$}
        & \textbf{Permutation $p$}
        & \boldmath$p_{\mathrm{FDR}}$ \\
        \midrule

        \multirow{2}{*}{CelebA~\cite{liu2015faceattributes}}
        & 10
        & -0.078
        & -0.174
        & 0.281
        & 0.561 \\

        & 1
        & -0.114
        & -0.055
        & 0.721
        & 0.721 \\

        \midrule

        \multirow{2}{*}{FairFace~\cite{karkkainen2021fairface}}
        & 10
        & -0.726
        & -0.674
        & 0.003
        & 0.006 \\

        & 1
        & 0.023
        & -0.143
        & 0.583
        & 0.583 \\

        \bottomrule
    \end{tabular}
    }
\end{table}

\subsection{Per-Attribute AUC Degradation}
Figure \ref{fig:auc_degradation} shows the relationship between positive attribute prevalence and AUC degradation under differential privacy. Statistical results are summarized in Table \ref{tab:auc_degradation_statistics}. For FairFace, prevalence is strongly negatively correlated with AUC degradation at $\epsilon=10$ (Pearson $r=-0.73$, Spearman $\rho=-0.67$, $p_{\mathrm{FDR}}=0.006$), indicating greater degradation among less prevalent groups. In contrast, CelebA shows weak correlations at both privacy budgets ($|\rho| \leq 0.17$, $p_{\mathrm{FDR}} \geq 0.56$), while the relationship for FairFace becomes weak at $\epsilon=1$ ($\rho=-0.14$, $p_{\mathrm{FDR}}=0.58$). These results suggest that attribute prevalence alone does not consistently explain AUC degradation across datasets and privacy budgets.

Several notable outliers further illustrate this heterogeneity. The CelebA attributes \textit{Smiling}, \textit{Wearing Necktie}, and \textit{Eyeglasses} exhibit substantial AUC degradation despite relatively high prevalence. Similarly, the well-represented FairFace groups 20--29 and Latino-Hispanic experience greater than 20\% degradation despite exceeding 15\% prevalence. One potential explanation is that attributes differ in the amount and spatial distribution of visual information required for classification. Several attributes with high observed degradation have relatively small spatial support, concentrating their visual cues in limited facial regions and potentially providing less redundant information to withstand DP noise. In contrast, attributes with more spatially distributed cues may be more robust. Other factors, including classification difficulty, intra-class variation, and dependencies among correlated attributes, may further contribute to this heterogeneity.

\section{Conclusion}
We investigated the interaction between differential privacy and fairness in generative models of human faces across multiple datasets and privacy levels. Our results show that stronger privacy can disproportionately degrade the quality and utility of some underrepresented groups, but this effect is not universal. In particular, attribute prevalence alone does not explain the observed degradation, with several well-represented attributes also experiencing substantial performance loss. These findings demonstrate that the relationship between privacy and fairness is more nuanced than representation alone and highlight the need for granular evaluation of demographic and semantic attributes when developing privacy-preserving generative models.

\clearpage
\bibliographystyle{IEEEbib}
\bibliography{ref}

\end{document}